# An Online Universal Classifier for Binary, Multi-class and Multi-label Classification

Meng Joo Er
School of EEE
Nanyang Technological University
Singapore
EMJER@ntu.edu.sg

Rajasekar Venkatesan
School of EEE
Nanyang Technological University
Singapore
RAJA0046@e.ntu.edu.sg

Ning Wang
Marine Engineering College
Dalian Maritime University
China
n.wang.dmu.cn@gmail.com

*Abstract*—Classification involves the learning of the mapping function that associates input samples to corresponding target label. There are two major categories of classification problems: Single-label classification and Multi-label classification. Traditional binary and multi-class classifications are sub-categories of single-label classification. Several classifiers are developed for binary, multi-class and multi-label classification problems, but there are no classifiers available in the literature capable of performing all three types of classification. In this paper, a novel online universal classifier capable of performing all the three types of classification is proposed. Being a high speed online classifier, the proposed technique can be applied to streaming data applications. The performance of the developed classifier is evaluated using datasets from binary, multi-class and multi-label problems. The results obtained are compared with state-of-the-art techniques from each of the classification types.

## I. Introduction

Machine learning classification is the process of approximating the mapping function that maps the input sample to target class/label [1]. In traditional classification problems, the input samples correspond to only one target label. This type of classification is called single-label classification. Binary classification involves classifying the input data samples into either of two sets based on a specific classification metric. The number of disjoint labels is 2 for binary classification. Disease diagnosis [2, 3], quality control, spam detection [4], malware detection are some of the major application areas of this method. But there are several real world application problems involving multiple target labels resulting in the development of multi-class classification. Multi-class classification involves classifying the input samples into more than two classes. Character recognition [5, 6], biometric identification [7] and security, face recognition are some of the application areas of multi-class classification. The single-label classification techniques are based on the assumption of unique target label association. i.e. each input sample corresponds to only one target label. In other words, the labels corresponding to the input samples form a disjoint set.

However, in many real world applications, the input samples correspond to multiple target labels. This condition of classification, where the input data correspond to a set of class labels instead of one, is called multi-label classification. Multi-label classification has become a rapidly emerging field of machine learning due to the wide range of application domains and the omnipresence of multi-label problems in real world scenarios [8]. The application areas of multi-label classification includes image, music and video categorization [9], medical diagnosis, bioinformatics, multimedia, genomics etc. In contrast to single-label classification, each sample may have multiple target labels in multi-label classification [10]. The generalization of multi-label classification results in the increased complexity of the classifier. Different classification techniques based on multi-layer perceptrons (MLP), decision trees (DT), k-nearest neighbors (kNN), support vector machine (SVM), extreme learning machine (ELM), naïve bayes classifier etc. have been developed for each of the classification types and is available in the literature [1, 11-14].

Based on the learning style, the machine learning techniques can be classified into batch learning and online learning techniques. In batch learning, the data required for the training of the classifier are collected in prior. The entire training data is processed concurrently for the estimation of the system parameters. The requirement to have all the training data in prior to training poses a serious constraint in the application of batch learning techniques. On the other hand, in online learning, the system parameters are updated in an iterative manner with sequential data. Therefore, online learning techniques are preferred over batch learning techniques for streaming data applications [15, 16].

There are several machine learning techniques available for binary, multi-class and multi-label classifications individually. There are no classifiers available in the literature that is capable of performing all three types of classification. In this paper, we propose an extreme learning machine based online universal classifier that is independent of classification type and can perform all three types of classification (binary, multi-class and multi-label). The developed classifier will identify both the classification type and the target labels associated with the input samples. The proposed classifier is experimented with datasets corresponding to each of the three classification types and is evaluated for consistency, speed and performance. The results are compared with the state-of-the-art techniques of the individual classification type.

The rest of the paper is organized as follows. Section 2 briefly summarizes the background information pertaining to the different classification types and on extreme learning machines. Section 3 describes the steps involved in the proposed approach. The details of the experimental design and the dataset specifications are discussed in Section 4. Section 5 summarizes the performance of the proposed classifier and the comparison of the experimental results with state-of-the-art techniques and concluding remarks are given in Section 6.

## II. BACKGROUND AND PRELIMINARIES

### A. Classification

Based on the label association to the input samples, the classification methods can be categorized into single-label classification and multi-label classification.

*1) Single-label Classification*

Single label classification is a function approximator that associates the input samples to a unique target label 'l' from a set of disjoint labels 'L'. The single-label classification problem can be further divided into two categories: Binary and multi-class classification [10]. When the input data samples are categorized into one of two classes, it is called binary classification. When the input samples correspond to one among a pool of target labels, it is called multi-class classification. Binary classification is the most basic classification and forms the basic requirement a technique should fulfill to be a classification method. Hence, all the classification methods and the variants available in the literature thus far can be used for binary classification. There are several methods existing in the literature to solve the multi-class classification problems. The existing multi-class classification methods can be classified into three groups.

- Extended methods from binary classification
- Decomposition to binary classification methods
- Hierarchical Classification methods

**Extended Methods from Binary Classification.** Some of the binary classification techniques can be directly extended to support the multi-class classification problems. Multi-class classification techniques based on multi-layer perceptron, decision trees, k-nearest neighbors, support vector machines, extreme learning machines and naïve bayes classifier are examples of algorithm adaptation methods.

**Decomposition to Binary Classification Methods.** The decomposition methods, as the name implies decompose the multi-class classification problem into multiple binary classification problems and employs the existing binary classifiers to solve it. Several methods have been proposed in the literature [11] that use decomposition to solve the multi-class classification problems [17, 18].

**Hierarchical Classification.** In hierarchical classification, the classes are arranged in the form of a tree. The parent node is divided to have leaf nodes such that each of the leaf node classes will be the subset of parent node classes. The similar procedure is extended until the leaf nodes have only one class label [19]. Realization of each of the nodes in the tree is performed using a binary classifier.

*2) Multi-label Classification*

Multi-label classification has gained much importance in recent years due to its wide range of application domains. As opposed to single-label classification, each input sample is associated with a set of target labels in multi-label classification. The number of target labels corresponding to each input is not fixed and varies dynamically. This results in increased complexity in the implementation of multi-label classifier [20]. Several methods have been developed for multi-label classification and is available in the literature. The existing techniques are grouped under three categories [10, 21].

- Algorithm Adaptation Methods
- Problem Transformation Methods
- Ensemble Methods

**Algorithm Adaptation Methods.** In the algorithm adaptation method, the base algorithm itself is extended to adapt for the multi-label classification. Several base algorithms have their multi-label variants such as Boosting, kNN, Decision Trees, Neural Networks, and SVM.

**Problem Transformation Methods.** In the problem transformation method, the multi-label classification is transformed to multiple binary or multi-class classification problems. Upon transforming the multi-label problem into multiple single-label problem, this technique utilizes the existing single-label classifiers to perform the classification and combining the results of all the single-label classifiers to find results for multi-label classification.

**Ensemble Methods.** Ensemble methods use an ensemble of algorithm adaptation and problem transformation methods and combine the results to perform multi-label classification.

The proposed method belongs to the category of algorithm adaptation method. The base algorithm itself is extended to adapt to all classification types.

### B. Extreme Learning Machine

Extreme learning machine (ELM) is proposed by Huang et al [22] has gained much attention due to its unique advantage of very high learning speed and random assignment of input weights. The universal approximation capability of single layer feedforward neural network is also preserved in ELM. Several variants of ELM have been developed and is available in the literature [23-28]. The proposed approach uses ELM based online universal classifier. A condensed overview of ELM is discussed below.

In ELM, the input weights of the neural network are randomly assigned. Therefore, only the output weights of the network are to be trained. Let N be the number of training samples and $\bar{N}$ be the number of hidden layer neurons. The output equation of ELM based network in matrix form is represented as

$$H\beta = Y \tag{1}$$

where H is the hidden layer output matrix of the network. The outputs of the hidden layer neurons corresponding to each input sample is populated as the column values of the H matrix. H is given by,

$$H = \begin{bmatrix} g(w_1 \cdot x_1 + b_1) & \cdots & g(w_N \cdot x_1 + b_N) \\ \vdots & \ddots & \vdots \\ g(w_1 \cdot x_N + b_1) & \cdots & g(w_N \cdot x_N + b_N) \end{bmatrix}_{N \times N} \quad (2)$$

$x_i = [x_{i1}, x_{i2}, \ldots, x_{in}]^T$ is the input data sample of dimension n, $g(x)$ is the activation function, $w_i = [w_{i1}, w_{i2}, \ldots w_{in}]^T$ is the input weight vector and b is the bias value of the network. β is the output weight matrix, Y is the target output corresponding to the input samples. During the training phase, the input sample and the output labels are given as inputs and the output weights of the ELM network is estimated using the equation

$$\beta = H^+ Y \quad (3)$$

$H^+ = (H^T H)^{-1} H^T$ gives the Moore-Penrose inverse of the H matrix. In the testing phase, the data samples are provided as the input and with the estimated β values, the target labels corresponding the input is predicted by the network. The mathematical background behind the functioning of the ELM has been extensively discussed in the literature [29, 30].

### III. PROPOSED APPROACH

An online universal classifier capable of performing classification on binary, multi-class and multi-label datasets is proposed. It is to be highlighted that there are no universal classifiers available in the literature that can classify all three classification types. Also, the proposed method is an online classifier and hence can be used for streaming data applications. The generality of the problem specification results in increased complexity in achieving universal classification technique. There are three key challenges to be addressed to achieve universal classifier.

1. Identification of classification type
2. Estimating the number of target labels corresponding to each input sample
3. Identifying each of the associated target labels.

The proposed approach is based on the online variant of ELM called online sequential extreme learning machine. The proposed approach falls under the category of an algorithm adaptation method in which the base algorithm is extended to adapt to the requirements of the universal classification. The various phases of the proposed algorithm are summarized.

**Initialization Phase.** Initialization Phase involves setting up the fundamental network parameters for the target classification problem. Being an ELM based technique, the input weights and the bias values are randomly initialized. The number of hidden layer neurons and the activation function are assigned. The number of hidden layer neurons is to be selected such that the problem of overfitting is avoided.

**Data Pre-processing Phase.** The proposed algorithm needs to be capable of classifying both single-label and multi-label classification problems. The representation of data varies among each of the classification types. In binary and multi-class classification, the output is represented as a single value which identifies the unique target class that is associated with the input sample. On the other hand, in multi-label classification, since each input can have multiple labels, the output is represented as a vector with dimensions equal to the total number of output labels. Thus, proper pre-processing of data is essential in achieving universal classifier. In the proposed approach, the target label of all three classification types is represented as a vector with dimension equal to the number of output labels. Each element in the vector signifies the belongingness of the input to the corresponding label.

**Online Training Phase.** During the training phase, the data samples and the target labels are provided as the input and the output weight values are estimated iteratively by online training. The proposed method is based on the online variant of ELM. The online training phase has two steps.

*Initial Block Step:* Let N0 be the number of training samples in the initial block of data, the initial output weight values are calculated using equations

$$M_0 = (H_0^T H_0)^{-1} \quad (4)$$

$$\beta_0 = M_0 H_0^T Y \quad (5)$$

*Sequential Training Step:* Upon completion of the initial block step, the subsequent data samples arriving sequentially are processed in the sequential training step. The output weight is updated iteratively with sequentially arriving data blocks using the recursive least square technique [24, 26]. The sequential update of output weight is given by the equations

$$M_{k+1} = M_k - \frac{M_k h_{k+1} h_{k+1}^T M_k}{1 + h_{k+1}^T M_k h_{k+1}} \quad (6)$$

$$\beta_{k+1} = \beta_k + M_{k+1} h_{k+1} (Y_{k+1}^T - h_{k+1}^T \beta_k) \quad (7)$$

By the end of the training phase, the values of β are estimated.

**Testing Phase.** In the testing phase, the target output of the input samples is predicted using the values of β estimated from the training phase and the input data samples. The raw output values of the network are evaluated using the relation Y = Hβ. The raw output value obtained from the testing phase is then processed to address the three challenges of the universal classifier.

**Classification Phase.** In the classification phase, the raw output values Y obtained from the training phase is used to predict the classification type, number of associated target labels and identifying each of the target labels corresponding to each input sample.

*Identifying the Classification Type:* The classification type of binary, multi-class or multi-label is identified using the classification type (CT) value and dimension of output vector 'l'. The CT value is evaluated using the equation.

$$CT = |HS(Y)| \quad (8)$$

where Y is the raw output vector and HS(x) is the heaviside function. Identification of classification type based on all possible valid combinations of CT and L is given in Table 1.

TABLE 1: IDENTIFICATION OF CLASSIFICATION TYPE

| CT = 1 | L = 2 | Binary Classification |
|---|---|---|
| CT = 1 | L > 2 | Multi-class Classification |
| CT > 1 | L > 2 | Multi-label Classification |

*Estimating the Number of Target Labels:* Upon establishing the classification type, the number of target labels is then estimated as given in Table 2. For binary and multi-class classification, the number of target labels is one, since each input belongs to unique target labels. For multi-label classification, the CT value corresponds to the number of target labels associated with the input data sample.

TABLE 2: ESTIMATION OF NUMBER OF TARGET LABELS

| Classification Type | Number of Target Labels |
|---|---|
| Binary Classification | 1 |
| Multi-class Classification | 1 |
| Multi-label Classification | $\sum_{i=1}^{L} HS(Y_i)$ |

*Identifying the Target Labels:* The target labels are identified using the belongingness vector. The belongingness vector B is given as,

$$B = HS(Y) \quad (9)$$

where Y is the raw output value and HS(x) is Heaviside function. Each element of the vector B denotes the belongingness of the input to the corresponding label. Thus, the label index of the non-zero entries of the B gives the target labels associated with the input samples. Upon estimating the target labels, the performance metrics of the classifier are evaluated. Thus, the proposed technique is capable of classifying all three types of classification problems. The overview of the proposed approach is summarized.

## IV. EXPERIMENTATION

The experimental design, dataset specifications and the comparison methods used to evaluate the proposed method are discussed in this section. Since the proposed method is capable of classifying all the classification types, the datasets from both single-label and multi-label classification problems are chosen for experimentation. In multi-label problems, different datasets have different degree of multi-labelness. Therefore, two metrics, Label cardinality and label density are used to quantitatively measure the degree of multi-labelness. Label cardinality gives the average number of labels corresponding to the input data. Therefore, for binary and multi-class classification, label cardinality will always be 1. Label density on the other hand also considers the number of labels in evaluation. Label cardinality and label density are very important metrics in the dataset specification for multi-label data. For example, label cardinality of 4.24 signifies that each of the input samples corresponds to more than 4 labels on an average. Two datasets having same label cardinality, but different label density can significantly vary the performance of the classifier. The specifications of the dataset used for experimentation are given in the Table 3.

**Algorithm: Proposed Universal Classifier Algorithm**
1. Initialization of parameters
2. Formatting input to uniform representation
3. Initial block training
   Input: Initial N0 samples of data in the form $\{(x_i,y_i)\}$
   Output: $\beta_0$
   Evaluation:
   $M_0 = (H_0^T H_0)^{-1}$
   $\beta_0 = M_0 H_0^T Y_0$
4. Sequential training
   Input: Sequentially arriving data blocks in the form $\{(x_i,y_i)\}$
   Output: $\beta_k$
   Evaluation:
   $$M_{k+1} = M_k - \frac{M_k h_{k+1} h_{k+1}^T M_k}{1 + h_{k+1}^T M_k h_{k+1}}$$
   $$\beta_{k+1} = \beta_k + M_{k+1} h_{k+1} (Y_{k+1}^T - h_{k+1}^T \beta_k)$$
5. Evaluating raw output value Y
   Input: Data sample xi
   Output: Y
   Evaluation:
   Y = Hβ
6. Evaluating CT value: $CT = |H(Y)|$
7. Identifying classification type based on CT and L
8. Evaluating number of associated target labels
9. Calculating the belongingness vector: B=H(Y)
10. Identifying the associated target labels
11. Evaluation of performance metrics corresponding to classification type

The proposed method is evaluated using 5 single-label (2 binary and 3 multi-class) datasets and 4 multi-label datasets. The datasets cover a wide range of feature dimension, number of labels, label density and label cardinality. The performance results of the proposed method on these datasets are compared with state-of-the-art techniques in the specific classification type. The performance of the proposed method on single-label classification datasets are compared with Support Vector Machine (SVM), kNN (k-Nearest Neighbor), MLP (Multi-layer Perceptron) and ELM (Extreme Learning Machine) based techniques. The results of the multi-label classification datasets are compared with the state-of-the-art methods based on SVM, kNN, DT (Decision Trees) and RF (Random Forest).

## V. RESULTS AND DISCUSSIONS

This section summarizes the experimental results of the proposed universal classifier on the datasets specified in table. Being an online method, the proposed algorithm can be used for streaming data applications.

### A. Consistency

Consistency is one of the key virtues of any new technique developed. An algorithm that is inconsistent with results on different trials is unreliable. Cross-validation is one of the effective ways to evaluate the consistency of the method. A 10-fold cross validation is performed for each of the datasets. In single-label classification problems, since each of the sample belongs to only one output, the performance of the classifier can be evaluated using the percentage of accuracy.

TABLE 3. DATASET SPECIFICATIONS

| Classification type | | Dataset | Number of labels | Feature dimension | Number of Samples | Label Cardinality | Label Density |
|---|---|---|---|---|---|---|---|
| Single-label | Binary | Diabetes | 2 | 8 | 768 | 1.00 | 0.500 |
| | | Ionosphere | 2 | 34 | 351 | 1.00 | 0.500 |
| | Multi-class | Iris | 3 | 4 | 150 | 1.00 | 0.333 |
| | | Waveform | 3 | 21 | 5000 | 1.00 | 0.333 |
| | | Balance-scale | 3 | 4 | 625 | 1.00 | 0.333 |
| Multi-label | | Scene | 6 | 294 | 2407 | 1.07 | 0.178 |
| | | Yeast | 14 | 103 | 2417 | 4.24 | 0.303 |
| | | Corel5k | 374 | 499 | 5000 | 3.53 | 0.009 |
| | | Enron | 53 | 1001 | 1702 | 3.38 | 0.064 |

However, multi-label classification poses a unique problem of partial correctness of the results. Therefore, a different set of performance metrics is used for evaluation. Hamming loss is one of the key performance metric for multi-label classification. It is the quantitative measure of the number of times the sample-label pair is misclassified. Lower the hamming loss, better the performance of the classifier. Hamming loss is evaluated as the summation of misclassified sample-label pair averaged over the total number of samples and labels. The performance of the binary and multi-class classifier is evaluated using percentage of accuracy and the performance of multi-label classifier is evaluated using hamming loss for the consistency evaluation. The results obtained are tabulated in Table 4. From the table, it can be seen that the proposed universal classifier is highly consistent for all datasets from binary, multi-class and multi-label classification.

*B. Speed*

The execution speed of the proposed classifier is evaluated in terms of training time and testing time. Execution speed of the classifier plays a vital role for streaming data applications. In order to perform real-time streaming data classification, the execution speed of the classifier should be less than the arrival rate of the streaming data. Therefore, a high speed classifier is essential for real-time streaming data applications. The proposed universal classifier exploits the inherent high-speed nature of the ELM. The training time and the testing time of the proposed universal algorithm for each dataset is given in Table 4. From the table, it is evident that the proposed classifier is capable of performing classification of all types with high speed, thus facilitating its application for real-time streaming data.

*C. Performance Comparison*

There are no universal classifier available in the literature to perform direct comparison with the proposed method. Therefore, the performance of the proposed classifier is compared with the state-of-the-art techniques in each of the classification type. For single-label classification datasets, the performance of the proposed method is compared with similar binary and multi-class techniques based on SVM, kNN, MLP and ELM. For multi-label classification problems, the performance of the proposed classifier is compared with SVM, kNN, DT and RF based techniques.

TABLE 4. CONSISTENCY AND SPEED

| Dataset | Accuracy % (10-fcv) | Training Time (s) | Testing Time (s) |
|---|---|---|---|
| Single-label Classification | | | |
| Diabetes | 78.2 ± 3.2 | 0.005 | 0 |
| Ionosphere | 96.4 ± 2.6 | 0.007 | 0 |
| Iris | 99.2 ± 0.6 | 0.002 | 0 |
| Waveform | 85.3 ± 1.8 | 0.012 | 0.001 |
| Balance-scale | 90.7 ± 3.7 | 0.009 | 0 |
| Multi-label Classification | | | |
| Scene | 0.096 ± 0.002 | 2.546 | 0.053 |
| Yeast | 0.201 ± 0.001 | 0.134 | 0.021 |
| Corel5k | 0.009 ± 0.000 | 5.521 | 0.079 |
| Enron | 0.047 ± 0.001 | 0.652 | 0.043 |

TABLE 5. PERFORMANCE COMPARISON

| Single-label Classification | | | | | |
|---|---|---|---|---|---|
| | Accuracy % | | | | |
| Dataset | SVM | kNN | MLP | ELM | Universal Classifier |
| Diabetes | 77.5 | 76.7 | 76.4 | 78.1 | 78.2 |
| Ionosphere | 94.9 | 96.7 | 96.0 | 96.6 | 96.4 |
| Iris | 98.7 | 98.4 | 99.2 | 98.6 | 99.2 |
| Waveform | 85.7 | 84.3 | 85.1 | 84.5 | 85.3 |
| Balance-scale | 91.2 | 90.3 | 88.6 | 89.3 | 90.7 |
| Multi-label Classification | | | | | |
| | Hamming loss | | | | |
| Dataset | SVM | kNN | RF-PCT | RFML-C4.5 | Universal Classifier |
| Scene | 0.082 | 0.099 | 0.094 | 0.116 | 0.096 |
| Yeast | 0.193 | 0.198 | 0.197 | 0.205 | 0.201 |
| Corel5k | 0.012 | 0.009 | 0.009 | 0.009 | 0.009 |
| Enron | 0.048 | 0.051 | 0.046 | 0.047 | 0.047 |

The results are given in the Table 5. Presence of the possibility of partial correctness in multi-label classification results in the need for evaluation of other parameters such as accuracy and F1 measure for multi-label classification. The results are tabulated in Table 6. From the comparison table it is evident that the proposed universal classifier performs uniformly well in datasets of all classification types.

TABLE 6. MULTI-LABEL PERFORMANCE METRICS

| Dataset | Accuracy | Precision | Recall | F1-measure |
|---------|----------|-----------|--------|------------|
| Scene   | 0.615    | 0.634     | 0.642  | 0.638      |
| Yeast   | 0.498    | 0.697     | 0.582  | 0.634      |
| Corel5k | 0.062    | 0.179     | 0.061  | 0.091      |
| Enron   | 0.408    | 0.645     | 0.464  | 0.540      |

## VI. CONCLUSIONS

A novel online universal classifier based on extreme learning machine is proposed. It is to be highlighted that there are no classifiers available in the literature that can classify binary, multi-class and multi-label classification. The proposed online universal classifier is experimented with nine different datasets of different classification types and the results are compared with state-of-the-art techniques in each type of classification problem. The proposed classifier is evaluated in terms of consistency, speed and performance. The high speed nature of the proposed classifier makes it suitable for real-time streaming data applications.

ACKNOWLEDGEMENT

The authors would like to acknowledge the funding support from the Ministry of Education, Singapore (Tier 1 AcRF, RG30/14), the National Natural Science Foundation of P. R. China (under Grants 51009017 and 51379002), Applied Basic Research Funds from Ministry of Transport of P. R. China (under Grant 2012-329-225-060), and Pro-gram for Liaoning Excellent Talents in University (under Grant LJQ2013055). Rajasekar Venkatesan is supported by NTU Research Student Scholarship.